\documentclass[10pt,twocolumn,letterpaper]{article}

\usepackage{wacv}
\usepackage{times}
\usepackage{epsfig}
\usepackage{graphicx}
\usepackage{microtype}
\usepackage{subfigure}
\usepackage[table,xcdraw]{xcolor}
\usepackage{booktabs} % for professional tables
\usepackage{amssymb,amsmath,bm}
\usepackage{mathtools}

\usepackage{algorithmic}
\usepackage{algorithm}

\def\wacvPaperID{****} % Enter the WACV Paper ID here

\wacvfinalcopy % *** Uncomment this line for the final submission

\ifwacvfinal
\fi

\ifwacvfinal
\usepackage[breaklinks=true,bookmarks=false]{hyperref}
\else
\usepackage[pagebackref=true,breaklinks=true,colorlinks,bookmarks=false]{hyperref}
\fi

\ifwacvfinal
\else
\fi

\begin{document}

%%%%%%%%% TITLE
\title{Adversarial Deepfakes: Evaluating Vulnerability of Deepfake Detectors to Adversarial Examples}

% \author{Shehzeen Hussain\\
% \and
% Paarth Neekhara\\

\author{\text{*}Shehzeen Hussain, \text{*}Paarth Neekhara, Malhar Jere, Farinaz Koushanfar, Julian McAuley\\
University of California San Diego\\
{\tt\footnotesize \{pneekhar,ssh028\}@ucsd.edu} \\
\text{*} Equal contribution\\
}

\maketitle
\thispagestyle{empty}

%%%%%%%%% ABSTRACT
\begin{abstract}
Recent advances in video manipulation techniques have made the generation of fake videos more accessible than ever before. Manipulated videos can fuel disinformation and reduce trust in media. Therefore detection of fake videos has garnered immense interest in academia and industry. Recently developed Deepfake detection methods rely on Deep Neural Networks (DNNs) to distinguish AI-generated fake videos from real videos. In this work, we demonstrate that it is possible to bypass such detectors by adversarially modifying fake videos synthesized using existing Deepfake generation methods. We further demonstrate that our adversarial perturbations are robust to image and video compression codecs, making them a real-world threat. We present pipelines in both white-box and black-box attack scenarios that can fool DNN based Deepfake detectors into classifying fake videos as real.
\end{abstract}

%%%%%%%%% BODY TEXT
\section{Introduction}

% What are deepfakes and why are they a threat.

With the advent of sophisticated image and video synthesis techniques, it has become increasingly easier to generate high-quality convincing fake videos.
Deepfakes are a new genre of synthetic videos, in which a subject's face is modified into a target face in order to simulate the target subject in a certain context and create convincingly realistic footage of events that never occurred. Video manipulation methods like Face2Face~\cite{thies2016face2face}, Neural Textures~\cite{thies2019deferred} and FaceSwap~\cite{faceswap} operate end-to-end on a source video and target face and require minimal human expertise to generate fake videos in real-time.

The intent of generating such videos can be harmless and have advanced the research of synthetic video generation for movies, storytelling and modern-day streaming services. However, they can also be used maliciously to spread disinformation, harass individuals or defame famous personalities~\cite{suwajanakorn2017synthesizing}. The extensive spread of fake videos through social media platforms 
has raised significant concerns worldwide, particularly hampering the credibility of digital media.
\begin{figure}[h]
    \centering
    \includegraphics[width=1.0\columnwidth]{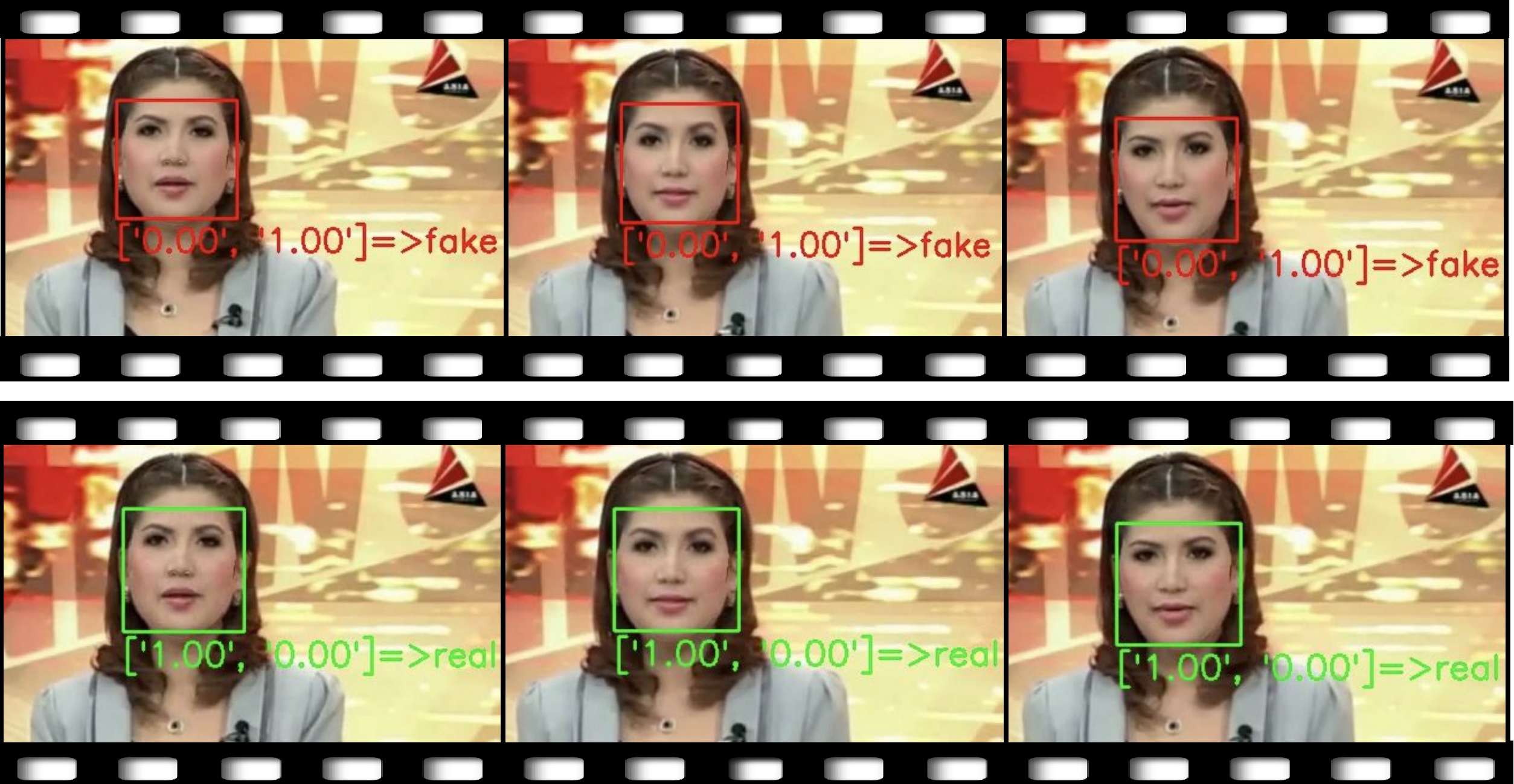}{\centering}
    \caption{Adversarial Deepfakes for XceptionNet~\cite{faceforensicsiccv} detector. Top: Frames of of a \textit{fake} video generated by Face2Face being correctly identified as \textit{fake} by the detector.
    Bottom: Corresponding frames of the adversarially modified fake video being classified as \text{real} by the detector.}
    \label{fig:intropicture}
\end{figure}

To address the threats imposed by Deepfakes, the machine learning community has proposed several countermeasures to identify forgeries in digital media. 
Recent state-of-the-art methods for detecting manipulated facial content in videos rely on Convolutional Neural Networks (CNNs)~\cite{dolhansky2019deepfake,faceforensicsiccv, afchar2018mesonet,  Amerini_2019_ICCV, li2019exposing,rahmouni2017distinguishing}. A typical 
Deepfake detector consists of a face-tracking method, following which the cropped face is passed on to a CNN-based classifier for classification as real or fake~\cite{afchar2018mesonet,chollet2017xception}. Some of the recent DeepFake detection methods use models operate on a sequence of frames as opposed to a single frame to exploit temporal dependencies in videos~\cite{sequencedetector}. 

While the above 
neural network based detectors achieve promising results in accurately detecting manipulated videos,
in this paper
we demonstrate that they are susceptible to \textit{adversarial examples} which can fool the detectors to classify fake videos as real \footnote{\scriptsize{Video Examples: }\url{https://adversarialdeepfakes.github.io/}}.
An adversarial example is an intentionally perturbed input that can fool a victim classification model~\cite{42503}. 
Even though several works have demonstrated that neural networks are vulnerable to adversarial inputs (Section \ref{sec:advexample}), we want to explicitly raise this issue that has been ignored by existing works on Deepfake detection (Section \ref{sec:detectors}). 
Since fake video generation can potentially be used for malicious purposes, it is critical to address the vulnerability of Deepfake detectors to adversarial inputs.

To this end, we quantitatively assess the vulnerability of state-of-the-art Deepfake detectors to adversarial examples. Our proposed methods can augment existing techniques for generating fake videos, such that they can bypass a given fake video detector. We generate adversarial examples for each frame of a given fake video and combine them together to synthesize an adversarially modified video that gets classified as real by the victim Deepfake detector.
We demonstrate that it is possible to construct fake videos that are robust to image and video compression codecs, making them a real world threat since videos shared over social media are usually compressed.  More alarmingly, we demonstrate that it is possible to craft robust adversarial Deepfakes in black-box settings, where the adversary may not be aware of the classification model used by the detector. Finally, we discuss normative points about how the community should approach the problem of Deepfake detection.

%-------------------------------------------------------------------------
\section{Background}
\subsection{Generating Manipulated Videos}
\label{sec:dfmethods}
Until recently, the ease of generating manipulated videos has been limited by manual editing tools. However, since the advent of deep learning and inexpensive computing services, there has been significant work in developing new techniques for automatic digital forgery. In our work, we generate adversarial examples for fake videos synthesized using FaceSwap (FS)~\cite{faceswap}, Face2Face (F2F)~\cite{thies2016face2face}, DeepFakes (DF)~\cite{DeepFakesgit} and NeuralTextures (NT)~\cite{thies2019deferred}. % the following methods:
We perform our experiments on this FaceForesics++ dataset~\cite{faceforensicsiccv}, which is a curated dataset of manipulated videos containing facial forgery using the above methods. Another recently proposed dataset containing videos with facial forgery is the DeepFake Detection Challenge (DFDC) Dataset~\cite{dolhansky2019deepfake}, which we utilize when evaluating our attacks against sequence based detection frameworks (Section ~\ref{sec:victimmodels}).

\subsection{Detecting Manipulated Videos}
\label{sec:detectors}
% Several earlier papers in
Traditionally, multimedia forensics investigated the authenticity of images~\cite{wang2007exposing,bohme2013digital,mitbook} using hand-engineered features and/or a-priori knowledge of the statistical and physical properties of natural photographs. 
However, video synthesis methods can be trained to bypass hand-engineered detectors by modifying their training objective. We direct readers to~\cite{barni2018adversarial,bohme2013counter} for an overview of counter-forensic attacks to bypass traditional (non-deep learning based) methods of detecting forgeries in multimedia content.

More recent works have employed CNN-based approaches that decompose videos into frames to automatically extract salient and discriminative visual features pertinent to Deepfakes. 
Some efforts have focused on segmenting the entire input image to detect facial tampering resulting from face swapping~\cite{66zhou2017two}, face morphing~\cite{raghavendra2017transferable} and splicing attacks~\cite{bappy2017exploiting, bappy2019hybrid}. Other works~\cite{li2020face,li2018ictu, afchar2018mesonet,guera2018deepfake,faceforensicsiccv,sabir2019recurrent} have focused on detecting face manipulation artifacts resulting from Deepfake generation methods. The authors of~\cite{li2018ictu} reported that eye blinking is not well reproduced in fake videos, and therefore proposed a temporal approach using a CNN + Recurrent Neural Network(RNN) based model to detect a lack of eye blinking when exposing deepfakes. Similarly,~\cite{yang2019exposing} used the inconsistency in head pose to detect fake videos. However, this form of detection can be circumvented by purposely incorporating images with closed eyes and a variety of head poses in training~\cite{vougioukas2019realistic,duarte2019wav2pix}. 

The Deepfake detectors proposed in~\cite{faceforensicsiccv, afchar2018mesonet,dolhansky2019deepfake} model Deepfake detection as a per-frame binary classification problem. The authors of~\cite{faceforensicsiccv} demonstrated that XceptionNet can outperform several alternative classifiers in detecting forgeries in both uncompressed and compressed videos, and identifying forged regions in them. In our work, we expose the vulnerability of such state-of-the-art Deepfake detectors. Since the task is to specifically detect facial manipulation, these models incorporate domain knowledge by using a face tracking method~\cite{thies2016face2face} to track the face in the video. The face is then cropped from the original frame and fed as input to classification model to be labelled as \textit{Real} or \textit{Fake}. Experimentally, the authors of~\cite{faceforensicsiccv} demonstrate that incorporation of domain knowledge helps improve classification accuracy as opposed to using the entire image as input to the classifier. The best performing classifiers amongst others studied by~\cite{faceforensicsiccv} were both CNN based models: XceptionNet~\cite{chollet2017xception} and MesoNet~\cite{afchar2018mesonet}.
More recently, some detectors have also focused on exploiting temporal dependencies while detecting DeepFake videos. Such detectors work on sequence of frames as opposed to a single frame using a CNN + RNN model or a 3-D CNN model. One such model based on a 3-D EfficientNet~\cite{tan2019efficientnet} architecture, was used by the third place winner~\cite{sequencedetector} of the recently conducted DeepFake Detection Challenge (DFDC)~\cite{dolhansky2019deepfake}. The first two winning submissions were CNN based per-frame classification  models similar to ones described above. We evaluate our attacks against the 3D CNN model to expose the vulnerability of temporal Deepfake detectors.

\subsection{Adversarial Examples}
\label{sec:advexample}
Adversarial examples are intentionally designed inputs to a machine learning (ML) model that cause the model to make a mistake \cite{42503}.  Prior work has shown a series of first-order gradient-based attacks to be fairly effective in fooling DNN based models in both image~\cite{limitations,papernot1,goodfellow2014explaining,Moosavi-Dezfooli_2017_CVPR,carlini2017towards,tramer,shi2019curls}, audio~\cite{targetattacks,yaoaudio,neekhara2019universal} and text~\cite{ebrahimi2018hotflip,belinkov2018synthetic,neekharatextadversarial} domains. The objective of such adversarial attacks is to find a good trajectory that (i)~maximally changes the value of the model's output and (ii)~pushes the sample towards a low-density region. This is equivalent to the ML model's gradient with respect to input features.
Prior work on defenses~\cite{xie2018mitigating} against adversarial attacks, propose to perform random operations over the input images, e.g., random cropping and JPEG compression. However, such defenses are shown to be vulnerable to attack algorithms that are aware of the randomization approach. Particularly, one line of adversarial attack~\cite{athalye2018obfuscated,eot} computes the expected value of gradients for each of the sub-sampled networks/inputs and performs attacks that are robust against compression. 
%-------------------------------------------------------------------------
\vspace{-2mm}
\section{Methodology}
\label{sec:methodology}
\subsection{Victim Models: Deepfake Detectors}
\label{sec:victimmodels}
\noindent \textbf{Frame-by-Frame detectors: }
To demonstrate the effectiveness of our attack on Deepfake detectors, we first choose detectors which rely on frame level CNN based classification models. These victim detectors work on the frame level and classify each frame independently as either \textit{Real} or \textit{Fake} using the following two-step pipeline:

\noindent \textbf{1.} A face tracking model~\cite{thies2016face2face} extracts the bounding box of the face in a given frame.\\
\noindent \textbf{2.} The cropped face is then resized appropriately and passed as input to a CNN based classifier to be labelled as either real or fake. 

In our work, we consider two victim CNN classifiers:  XceptionNet~\cite{chollet2017xception} and MesoNet~\cite{afchar2018mesonet}.
Detectors based on the above pipeline have been shown to achieve state-of-the-art performance in Deepfake detection as reported in~\cite{dolhansky2019deepfake, faceforensicsiccv,7553523}. The accuracy of such models on the FaceForensics++ Dataset~\cite{faceforensicsiccv} is reported in Table~\ref{tab:detectoracc}.

\noindent \textbf{Sequence based models: }
We also demonstrate the effectiveness of our attacks on detectors that utilize temporal dependencies. 
Such detection methods typically use a CNN + RNN or a 3D-CNN architecture to classify a \textit{sequence} of frames as opposed to a single frame. A 3D-CNN architecture performs convolutions across height, width and time axis thereby exploiting temporal dependencies. 
In Section~\ref{sec:temporalevaluation}, we evaluate our attacks against one such detection method~\cite{sequencedetector} that uses a 3-D EfficientNet~\cite{tan2019efficientnet} CNN model for classifying a sequence of face-crops obtained from a face tracking model. In this model, a 3-D convolution is added to each block of the EfficientNet model to perform convolutions across time. The length of the input sequence to the model is 7 frames and the step between frames is 1/15 of a second. This 3-D CNN model was used by the third place winner of the recently conducted DFDC challenge.

\subsection{Threat Model}
\begin{figure}[t]
    \centering
    \includegraphics[width=0.8\linewidth]{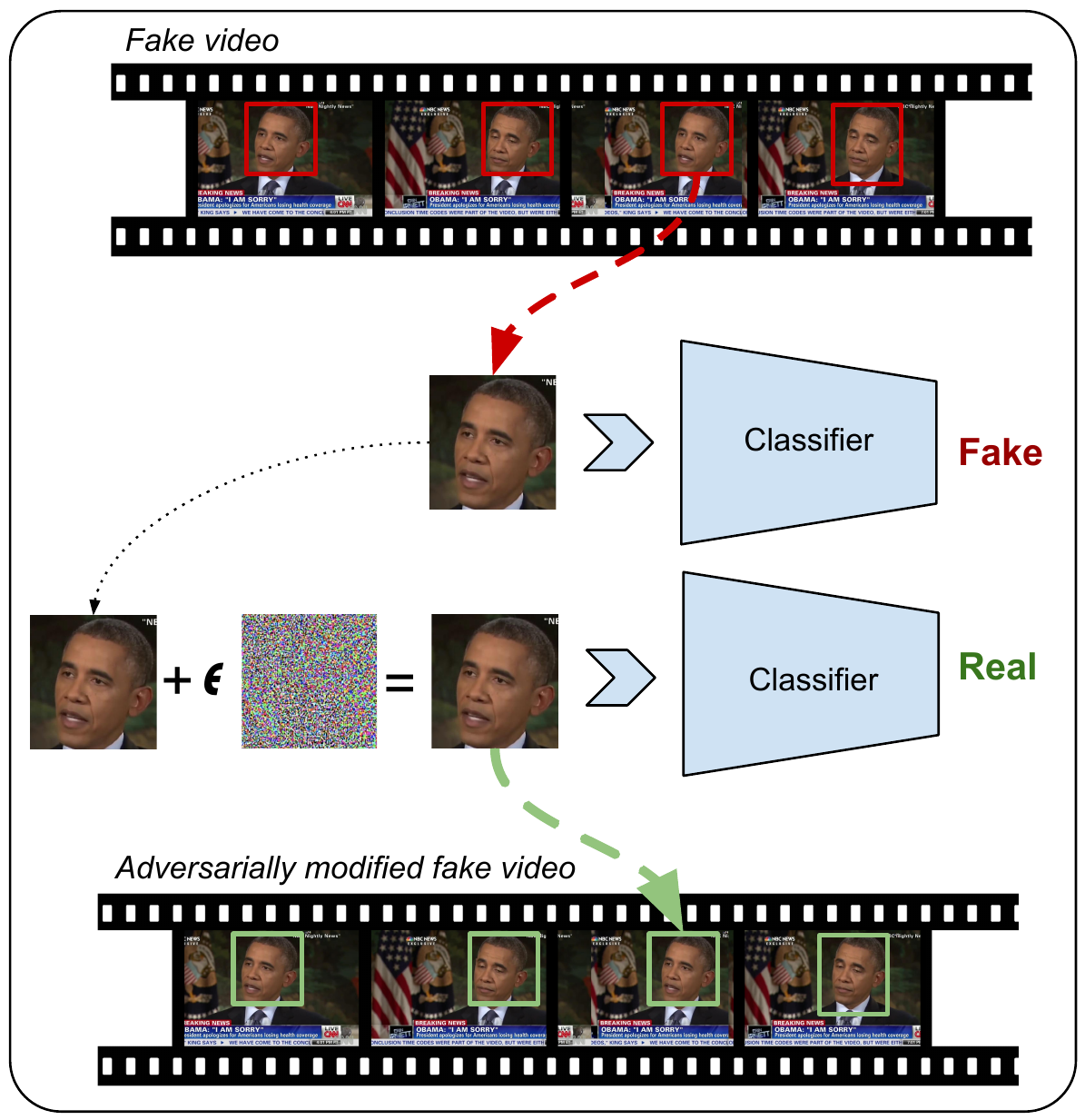}
    \caption{An overview of our attack pipeline to generate Adversarial Deepfakes. We generate an adversarial example for each frame in the given fake video and combine them together to create an adversarially modified fake video.}
    \label{fig:attack}
    \vspace{-2mm}
\end{figure}
Given a facially manipulated (fake) video input and a victim Deepfake detector, our task is to adversarially modify the fake video such that most of the frames get classified as \textit{Real} by the Deepfake detector, while ensuring that the adversarial modification is quasi-imperceptible.

\textbf{Distortion Metric:} To ensure imperceptibility of the adversarial modification, the $L_p$ norm is a widely used distance metric for measuring the distortion between the adversarial and original inputs. The authors of~\cite{goodfellow2014explaining} recommend constraining the maximum distortion of any individual pixel by a given threshold $\epsilon$,  i.e.,~constraining the perturbation using an $L_\infty$ metric. Additionally, \textit{Fast Gradient Sign Method} (FGSM)~\cite{goodfellow2014explaining} based attacks, which are optimized for the $L_\infty$ metric, are more time-efficient than attacks which optimize for $L_2$ or $L_0$ metrics. Since each video can be composed of thousands of individual frames, time-efficiency becomes an important consideration to ensure the proposed attack can be reliably used in practice. Therefore, in this work, we use the $L_\infty$ distortion metric for constraining our adversarial perturbation and optimize for it using gradient sign based methods. 

\textbf{Notation:} We follow the notation previously used in~\cite{carlini2017towards,papernot2016distillation}: Define $F$ to be the full neural network (classifier) including the softmax function, $Z(x) = z$ to be the output of all layers except the softmax (so $z$ are the logits), and 
$$F(x) = softmax(Z(x)) = y$$
The classifier assigns the label $C(x) = \arg\max_i(F(x)_i)$ to input $x$.

\textbf{Problem Formulation:} Mathematically, for any given frame $x_0$ of a fake video, and a victim frame-forgery detector model $C$, we aim to find an adversarial frame $x_{adv}$ such that,
$$C(x_{\mathit{adv}}) = \mathit{Real} \text{ and } ||x_{\mathit{adv}}-x_0||_\infty < \epsilon$$ 

\textbf{Attack Pipeline:} An overview of the process of generating adversarial fake videos is depicted in Figure~\ref{fig:attack}. For any given frame, we craft an adversarial example for the cropped face, such that after going through some image transformations (normalization and resizing), it gets classified as $\textit{Real}$ by the classifier. The adversarial face is then placed in the bounding box of face-crop in the original frame, and the process is repeated for all frames of the video to create an adversarially modified fake video. In the following sections, we consider our attack pipeline under various settings and goals. 

Note that, the proposed attacks can also be applied on detectors that operate on entire frames as opposed to face-crops. We choose face-crop based victim models because they have been shown to outperform detectors that operate on entire frames for detecting facial-forgeries. 

\subsection{White-box Attack}
\label{sec:whitebox}
In this setting, we assume that the attacker has complete access to the detector model, including the face extraction pipeline and the architecture and parameters of the classification model. To construct adversarial examples using the attack pipeline described above, we use the iterative gradient sign method~\cite{kurakin2016adversarial} to optimize the following loss function:
\begin{equation}
\begin{split}
& \text{Minimize } \mathit{loss}(x') \text{ where}\\
& \mathit{loss}(x') = \mathit{max}(Z(x')_{\mathit{Fake}} - Z(x')_{\mathit{Real}}, 0) \\
\end{split}
\label{eq1}
\end{equation}

Here, $Z(x)_y$ is the final score for label $y$ before the softmax operation in the classifier $C$. Minimizing the above loss function maximizes the score for our target label $\mathit{Real}$. The loss function we use is recommended 
in~\cite{carlini2017towards} because it is empirically found to generate less distorted adversarial samples and is robust against defensive distillation. We use the iterative gradient sign method to optimize the above objective while constraining the magnitude of the perturbation as follows: 
\begin{equation}
\begin{split}
% & x'_{0} = x_{0} \\
& x_i = x_{i-1} - \text{clip}_{\epsilon}(\alpha \cdot \text{sign}(\nabla \mathit{loss}(x_{i-1})))\\
\end{split}
\label{eq2}
\end{equation}
We continue gradient descent iterations  until success or until a given number number of maximum iterations, whichever occurs earlier. In our experiments, we demonstrate that while we are able to achieve an average attack success rate of 99.05\% when we save videos with uncompressed frames, the perturbation is not robust against video compression codecs like \textit{MJPEG}.  
In the following section, we discuss our approach to overcome this limitation of our attack. 

\subsection{Robust White-box Attack}
\label{sec:robustwhitebox}
Generally, videos uploaded to social networks and other media sharing websites are compressed. Standard operations like compression and resizing are known for removing adversarial perturbations from an image~\cite{dziugaite2016study,das2017keeping,guo2017countering}. To ensure that the adversarial videos remain effective even after compression, we craft adversarial examples that are robust over a given distribution of input transformations~\cite{eot}. Given a distribution of input transformations $T$, input image $x$, and target class $y$, our objective is as follows:
$$ x_{adv} = \mathit{argmax}_{x} \mathbb{E}_{t\sim T} [F(t(x))_y] \text{ s.t. } ||x - x_0||_\infty < \epsilon $$
That is, we want to maximize the expected probability of target class $y$ over the distribution of input transforms $T$. To solve the above problem, we update the loss function given in Equation~\ref{eq1} to be an expectation over input transforms $T$ as follows:
$$\mathit{loss}(x) = \mathbb{E}_{t\sim T} [\mathit{max}(Z(t(x))_{\mathit{Fake}} - Z(t(x))_{\mathit{Real}}, 0)]$$
Following the law of large numbers, we estimate the above loss functions for $n$ samples as:
$$\mathit{loss}(x) = \frac{1}{n}\sum_{t_i\sim T} [\mathit{max}(Z(t_i(x))_{\mathit{Fake}} - Z(t_i(x))_{\mathit{Real}}, 0)]$$

Since the above loss function is a sum of differentiable functions, it is tractable to compute the gradient of the loss w.r.t.~to the input $x$. We minimize this loss using the iterative gradient sign method given by Equation~\ref{eq2}. We iterate until a given a number number of maximum iterations or until the attack is successful under the sampled set of transformation functions, whichever happens first.

Next we describe the class of input transformation functions we consider for the distribution $T$:\\
\textbf{Gaussian Blur:}  Convolution of the original image with a Gaussian kernel $\mathit{k}$. This transform is given by $t(x)= k \ast x $ where $\ast$ is the convolution operator.\\
\textbf{Gaussian Noise Addition:} Addition of Gaussian noise sampled from $\Theta\sim{\mathcal{N}(0,\sigma)}$ to the input image. This transform is given by $t(x)= x + \Theta $ \\
\textbf{Translation:} We pad the image on all four sides by zeros and shift the pixels horizontally and vertically by a given amount.
Let $t_x$ be the transform in the $x$ axis and $t_y$ be the transform in the $y$ axis, then $t(x) = x'_{H,W,C}$ s.t $x'[i,j,c] = x[i+t_x,j+t_y,c]$
\\
\textbf{Downsizing and Upsizing:} The image is first downsized by a factor $r$ and then up-sampled by the same factor using bilinear re-sampling.

The details of the hyper-parameter search distribution used for these transforms can be found in the Section~\ref{sec:whiteboxexp}. 
% Empirically, we find that ensuring robustness of adversarial examples to these transforms ensures robustness to JPEG compression. Intuitively, image translation transform ensures robustness to different crops of the face, while gaussian blur, gaussian noise and resizing ensure robustness to JPEG artifacts. 

\subsection{Black-box Attack}
\label{sec:blackbox}
In the black-box setting, we consider the more challenging threat model in which the adversary does not have access to the classification network architecture and parameters. We assume that the attacker has knowledge of the detection pipeline structure and the face tracking model. However, the attacker can solely query the classification model as a black-box function to obtain the probabilities of the frame being \textit{Real} or \textit{Fake}. Hence there 
is
a need to estimate the gradient of the loss function by querying the model and observing the change in output for different inputs, since we cannot backpropagate through the network. 

We base our algorithm for efficiently estimating the gradient from queries on the Natural Evolutionary Strategies (NES) approach of~\cite{nes,ilyas2018black}. 
Since we do not have access to the pre-softmax outputs $Z$, we aim to maximize the class probability $F(x)_y$ of the target class $y$. Rather than maximizing the objective function directly, NES maximizes the expected value of the function under a search distribution  $\pi(\theta|x)$. That is, our objective is:
$$\textit{Maximize: } \mathbb{E}_{\pi(\theta|x)}[F(\theta)_y]$$
This allows efficient gradient estimation in fewer queries as compared to finite-difference methods. From ~\cite{nes}, we know the gradient of expectation can be derived as follows:
\begin{align*}
    \nabla_x \mathbb{E}_{\pi(\theta|x)}\left[F(\theta)_y\right] &= \mathbb{E}_{\pi(\theta|x)}\left[F(\theta)_y\nabla_x \log\left(\pi(\theta|x)\right)\right] \\
\end{align*}
Similar to~\cite{ilyas2018black,nes}, we choose a search distribution $\pi(\theta|x)$ of random Gaussian noise around the current image $x$. That is, $\theta = x + \sigma \delta$ where $\delta \sim \mathcal{N}(0, I)$. 
Estimating the gradient with a population of $n$ samples yields the following variance reduced gradient estimate:
$$\nabla\mathbb{E}[F(\theta)] \approx \frac{1}{\sigma n}\sum_{i=1}^n \delta_i
F(\theta + \sigma\delta_i)_y$$

We use antithetic sampling to generate $\delta_i$ similar to ~\cite{salimans2017evolution,ilyas2018black}. That is, instead of generating $n$ values $\delta \sim \mathcal{N}(0, I)$, we sample Gaussian noise for $i \in \{1, \ldots,
\frac{n}{2}\}$ and set $\delta_j = -\delta_{n-j+1}$ for $j \in
\{(\frac{n}{2}+1), \ldots, n\}$. This optimization has been empirically shown to improve performance of NES. Algorthim~\ref{alg:nes2} details our implementation of estimating gradients using NES. The transformation distribution $T$ in the algorithm just contains an identity function i.e.,~$T = \{ I(x) \}$ for the black-box attack described in this section.

After estimating the gradient, we move the input in the direction of this gradient using iterative gradient sign updates to increase the probability of target class:
\begin{equation}
\begin{split}
% & x'_{0} = x_{0} \\
& x_i = x_{i-1} + \text{clip}_{\epsilon}(\alpha \cdot \text{sign}(\nabla F(x_{i-1})_y))\\
\end{split}
\label{eq3}
\end{equation}

\subsection{Robust Black-box Attack}
\label{sec:robustblackbox}
To ensure robustness of adversarial videos to compression, we incorporate Expectation over Transforms (Section~\ref{sec:robustwhitebox}) method in the black-box setting for constructing adversarial videos. 

To craft adversarial examples that are robust under a given set of input transformations $T$, we maximize the expected value of the function under a search distribution  $\pi(\theta|x)$ and our distribution of input transforms $T$. That is, our objective is to maximize:
$$\mathbb{E}_{t \sim T}[\mathbb{E}_{\pi(\theta|x)}\left[F(t(\theta))_y\right]]$$
Following the derivation in the previous section, the gradient of the above expectation can be estimated using a population of size $n$ by iterative sampling of $t_i$ and $\delta_i$:
$$\nabla\mathbb{E}[F(\theta)] \approx \frac{1}{\sigma n}\sum_{i=1, t_i \sim T}^n \delta_i
F(t_i(\theta + \sigma\delta_i))_y$$

\begin{algorithm}
   \caption{NES Gradient Estimate}
   \label{alg:nes2}
\begin{algorithmic}
    \STATE {\bfseries Input:} Classifier $F(x)$,target class $y$, image $x$
    \STATE {\bfseries Output:} Estimate of $\nabla_x F(x)_y$
    \STATE {\bfseries Parameters:} Search variance $\sigma$, number of
    samples $n$, image dimensionality $N$
   \STATE $g \gets \bm{0}_n$
   \FOR{$i=1$ {\bfseries to} $n$}
%   \FOR{$t \text{ in }T$}
    \STATE $t_i \sim T$
	\STATE $u_i \gets \mathcal{N}(\bm{0}_{N}, \bm{I}_{N\cdot N})$
	\STATE $g \gets g + F(t_i(x+\sigma\cdot u_i))_y\cdot u_i$
	\STATE $g \gets g - F(t_i(x-\sigma\cdot u_i))_y\cdot u_i$
    % \ENDFOR
   \ENDFOR
   \STATE {\bfseries return} $\frac{1}{2n\sigma} g$
\end{algorithmic}
\end{algorithm}

We use the same class of transformation functions listed in Section~\ref{sec:robustwhitebox} for the distribution $T$. 
Algorithm~\ref{alg:nes2} details our implementation for estimating gradients for crafting robust adversarial examples. We follow the same update rule given by Equation~\ref{eq3} to generate adversarial frames. We iterate until a given a number of maximum iterations or until the attack is successful under the sampled set of transformation functions.

\label{sec:blackbox2}

%-------------------------------------------------------------------------

\section{Experiments}

\textbf{Dataset and Models: }We evaluate our proposed attack algorithm on two pre-trained victim models: XceptionNet~\cite{chollet2017xception} and MesoNet~\cite{afchar2018mesonet}. 
In our experiments, we perform our attack on the test set of the FaceForensics++ Dataset~\cite{faceforensicsiccv}, consisting of manipulated videos from the four methods described in Section~\ref{sec:dfmethods}.
We construct adversarially modified fake videos on the FaceForensics++ test set, which contains 70 videos (total 29,764 frames) from each of the four manipulation techniques. For simplicity, our experiments are performed on high quality (HQ) videos, which apply a light compression on raw videos.
The accuracy of the detector models for detecting facially manipulated videos on this test set is reported in Table~\ref{tab:detectoracc}.
% We will be releasing code for all our attack algorithms in PyTorch\footnote{Code released upon publication}.

\begin{table}[h]
\centering
\begin{tabular}{c|c|c|c|c|}
 \cline{2-5}
 & \textbf{DF} & \textbf{F2F} & \textbf{FS} & \textbf{NT} \\ \hline
\multicolumn{1}{|c|}{\textbf{XceptionNet Acc \%}} & 97.49 & 97.69 & 96.79 & 92.19 \\ \hline
\multicolumn{1}{|c|}{\textbf{MesoNet Acc \%}} & 89.55 & 88.6 & 81.24 & 76.62 \\ \hline
\multicolumn{1}{c}{}
\end{tabular}
\caption{Accuracy of Deepfake detectors on the FaceForensics++ HQ Dataset as reported in~\cite{faceforensicsiccv}. The results are for the entire high-quality compressed test set generated using four manipulation techniques (DF: DeepFakes, F2F: Face2Face, FS: FaceSwap and NT: NeuralTextures).}
\label{tab:detectoracc}
\end{table}

\textbf{Evaluation Metrics: }Once the adversarial frames are generated, we combine them and save the adversarial videos in the following formats:\\
1) \textit{ Uncompressed (Raw):} Video is stored as a sequence of uncompressed images.\\
2)\textit{ Compressed (MJPEG):} Video is saved as a sequence of JPEG compressed frames.\\
3)\textit{ Compressed (H.264):} Video is saved in the commonly used mp4 format that applies temporal compression across frames. 

We conduct our primary evaluation on the \textit{Raw} and \textit{MJPEG} video formats across all attacks. We also study the effectiveness of our white box robust attack using different compression levels in the \textit{H264} codec. We report the following metrics for evaluating our
attacks:
\\
\textbf{Success Rate (SR)}: The percentage of frames in the adversarial videos that get classified to our target label \textit{Real}. We report: \textbf{SR-U}- Attack success rate on uncompressed adversarial videos saved in Raw format; and \textbf{SR-C}-  Attack success rate on compressed  adversarial videos saved in MJPEG format.
\\
\textbf{Accuracy}: The percentage of frames in videos that get classified to their original label \textit{Fake} by the detector. We report \textbf{Acc-C}- accuracy of the detector on compressed adversarial videos. 
\\
\textbf{Mean distortion ($\mathbf{L_\infty}$)}: The average $L_\infty$ distortion between the adversarial and original frames. The pixel values are scaled in the range [0,1], so changing a pixel from full-on to full-off in a grayscale image would result in $L_\infty$ distortion of 1 (not 255). 

\subsection{White-box Setting}
\label{sec:whiteboxexp}
To craft adversarial examples in the white-box setting, in our attack pipeline, we implement differentiable image pre-processing (resizing and normalization) layers for the CNN. This allows us to backpropagate gradients all the way to the cropped face in-order to generate the adversarial image that can be placed back in the frame. We set the maximum number of iterations to $100$, learning rate $\alpha$ to $1/255$ and max $L_\infty$ constraint $\epsilon$ to $16/255$ for both our attack methods described in Sections \ref{sec:whitebox} and \ref{sec:robustwhitebox}.
\setlength\tabcolsep{2pt} % default value: 6pt
\begin{table}[h]
\centering
\resizebox{\columnwidth}{!}{%
\begin{tabular}{@{}c|cccc|cccc@{}}
  \multicolumn{1}{c}{} & \multicolumn{4}{c}{\emph{XceptionNet}} & \multicolumn{4}{c}{\emph{MesoNet}} \\ \midrule
\textbf{Dataset} & $\mathbf{L_\infty}$ & \textbf{SR - U} & \textbf{SR - C} & \textbf{Acc-C\%} & $\mathbf{L_\infty}$ & \textbf{SR - U} & \textbf{SR - C} & \textbf{Acc-C\%} \\ \midrule
\textbf{DF} & 0.004 & 99.67 & 43.11 & 56.89 & 0.006 & 97.30 & 92.27 & 7.73\\
\textbf{F2F} & 0.004 & 99.85 & 52.50 & 47.50 & 0.007 & 98.94 & 96.30 & 4.70 \\
\textbf{FS} & 0.004 & 100.00 & 43.13 & 56.87 & 0.009 & 97.12 & 86.10 & 13.90\\
\textbf{NT} & 0.004 & 99.89 & 95.10 & 4.90 & 0.007 & 99.22 & 96.20 & 3.80\\
\midrule
\textbf{All} & 0.004 & 99.85 & 58.46 & 41.54 & 0.007 & 98.15 & 92.72 & 7.28 \\ 
\bottomrule
\multicolumn{1}{c}{}
\end{tabular}%
}
\caption{Success Rate of White-box attack on XceptionNet and MesoNet. We report the average $L_\infty$ distortion between the adversarial and original frames and the attack success rate on uncompressed (SR-U) and compressed (SR-C) videos. Acc-C denotes the accuracy of the detector on compressed adversarial videos. }
\vspace{-4mm}
\label{tab:whitebox}
\end{table}

Table~\ref{tab:whitebox} shows the results of the white-box attack (Section~\ref{sec:whitebox}). We are able to generate adversarial videos with an average success rate of 99.85\% for fooling XceptionNet and 98.15\% for MesoNet when adversarial videos are saved in the Raw format. However, the attack average success rate drops to 58.46\% for XceptionNet and 92.72\% for MesoNet when MJPEG compression is used. This result is coherent with past works~\cite{dziugaite2016study,das2017keeping,guo2017countering} that employ JPEG compression and image transformations to defend against adversarial examples. 

\begin{figure}[htbp]
    \centering
    \includegraphics[width=1.0\columnwidth]{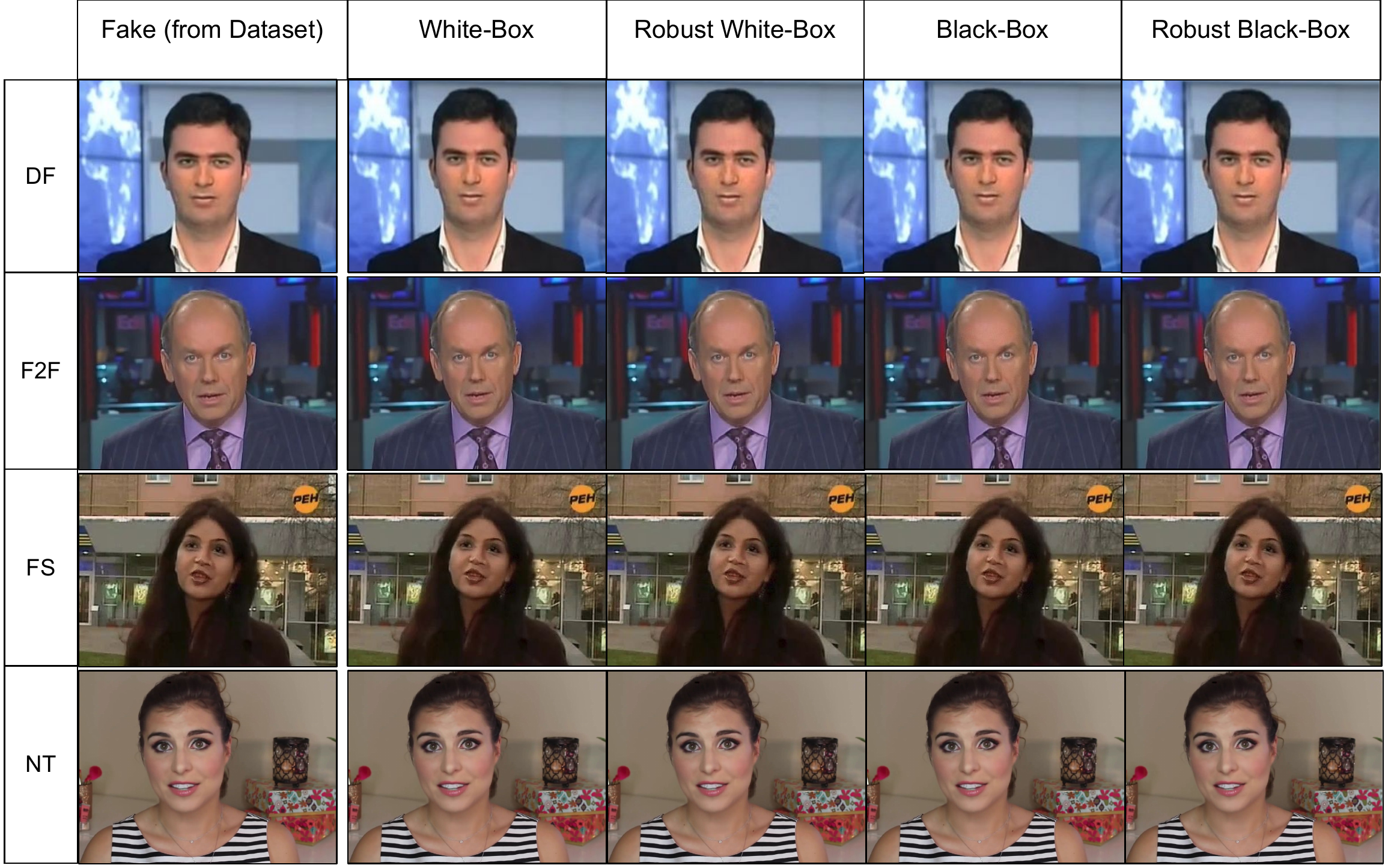}
    \caption{Randomly selected frames of Adversarial Deepfakes from successful attacks. The frame from the dataset in the first column is correctly identified as \textit{Fake} by the detectors, while the corresponding frames generated by each of our attacks are labelled as \textit{Real} with a probability of 1. Video examples are linked in the footnote on the first page.}
    \label{fig:advfakes}
\end{figure}

\textbf{Robust White-Box:} For our robust white box attack, we sample 12 transformation functions from the distribution $T$ for estimating the gradient in each iteration. This includes three functions from each of the four transformations listed in Section~\ref{sec:robustwhitebox}. Table~\ref{tab:transforms} shows the search distribution for different hyper-parameters of the transformation functions. 

\begin{table}[h]
\centering
\resizebox{\columnwidth}{!}{%
\begin{tabular}{@{}c|c@{}}
\toprule
\textbf{Transform} &  \textbf{Hyper-parameter search distribution}\\ 
\midrule
\textbf{Gaussian Blur} & Kernel $k(d,d,\sigma)$, $d \sim \mathcal{U}[3,7] $, $\sigma \sim \mathcal{U}[5,10] $\\
\textbf{Gaussian Noise} & $\sigma \sim \mathcal{U}[0.01, 0.02] $\\
\textbf{Translation} & $d_x \sim \mathcal{U}[-20, 20]$, $d_y \sim \mathcal{U}[-20, 20]$\\
\textbf{Down-sizing \& Up-sizing} & Scaling factor $r \sim \mathcal{U}[2, 5] $\\
\bottomrule
\multicolumn{1}{c}{}
\end{tabular}%
}
\caption{Search distribution of hyper-parameters of different transformations used for our Robust White box attack. During training, we sample three functions from each of the transforms to estimate the gradient of our expectation over transforms.}
\label{tab:transforms}
\vspace{-4mm}
\end{table}

\begin{table}[h]
\centering
\resizebox{\columnwidth}{!}{%
\begin{tabular}{@{}c|cccc|cccc@{}}
  \multicolumn{1}{c}{} & \multicolumn{4}{c}{\emph{XceptionNet}} & \multicolumn{4}{c}{\emph{MesoNet}} \\ \midrule
\textbf{Dataset} & $\mathbf{L_\infty}$ & \textbf{SR - U} & \textbf{SR - C} & \textbf{Acc-C\%} & $\mathbf{L_\infty}$ & \textbf{SR - U} & \textbf{SR - C} & \textbf{Acc-C\%} \\ \midrule
\textbf{DF} & 0.016 & 99.67 & 98.71 & 1.29 & 0.030 & 99.94 & 99.85 & 0.15  \\
\textbf{F2F} & 0.013 & 100.00 & 99.00 & 1.00 & 0.020 & 99.71 & 99.67 & 0.33  \\
\textbf{FS} & 0.013 & 100.00 & 95.33 & 4.67 & 0.026 & 99.02 & 98.50 & 1.50  \\
\textbf{NT} & 0.011 & 100.00 & 99.89 & 0.11 & 0.025 & 99.99 & 99.98 & 0.02  \\
\midrule
\textbf{All} & 0.013 & 99.91 & 98.23 & 1.77 & 0.025 & 99.67 & 99.50 & 0.50 \\
\bottomrule
\multicolumn{1}{c}{}
\end{tabular}%
}
\caption{Success Rate of Robust White-box attack on XceptionNet and MesoNet. Acc-C denotes the accuracy of the detector on compressed adversarial videos.}
\label{tab:whiteboxrobust}
\vspace{-2mm}
\end{table}

Table~\ref{tab:whiteboxrobust} shows the results of our robust white-box attack. It can be seen that robust white-box is effective in both Raw and MJPEG formats. The average distortion between original and adversarial frames in the robust attack is higher as compared to the non-robust white-box attack. We achieve an average success rate (SR-C) of 98.07\% and 99.83\% for XceptionNet and MesoNet respectively in the compressed video format.
Additionally, to assess the gain obtained by incorporating the transformation functions, we compare the robust white-box attack against the non-robust white-box attack at the same level of distortion in Table~\ref{tab:perturbcomparison}. We observe a significant improvement  in attack success rate on compressed videos (SR-C) when using the robust attack as opposed to the simple white-box attack (84.96\% vs 74.69\% across all datasets at $L_\infty$ norm of 0.008). 

\begin{table}[h]
\centering
\resizebox{\columnwidth}{!}{%
\begin{tabular}{@{}c|c|ccc|ccc@{}}
  \multicolumn{2}{c}{} & \multicolumn{3}{c}{\emph{White Box}} &
  \multicolumn{3}{c}{\emph{Robust White Box}} \\\midrule
\textbf{Dataset} & $\mathbf{L_\infty}$ & \textbf{SR - U} & \textbf{SR - C} &  \textbf{Acc-C\%} & \textbf{SR - U} & \textbf{SR - C} &  \textbf{Acc-C\%} \\ \midrule
\textbf{DF} & 0.008 & 99.67 & 60.36 & 39.64 & 99.67 & 75.06 & 24.94  \\
\textbf{F2F} & 0.008 & 99.85 & 80.69 & 19.31 & 100.0 & 90.20 & 9.80  \\
\textbf{FS} & 0.008 & 100.00 & 59.63 & 40.37 & 100.0 & 76.12 & 23.88  \\
\textbf{NT} & 0.008 & 99.89 & 98.08 & 1.92 & 100.0 & 98.48 & 1.52  \\
\midrule
\textbf{All} & 0.008 & 99.85 & 74.69 & 25.31 & 99.91 & 84.96 & 15.04  \\
\bottomrule
\multicolumn{1}{c}{}
\end{tabular}
}
\caption{Comparison of white-box and robust white-box attacks at the same magnitude of $L_\infty$ norm of the adversarial perturbation. Acc-C denotes the accuracy of the detector on compressed adversarial videos.}
\label{tab:perturbcomparison}
\end{table}

\begin{figure}[t]
    \centering
    \includegraphics[width=0.9\linewidth]{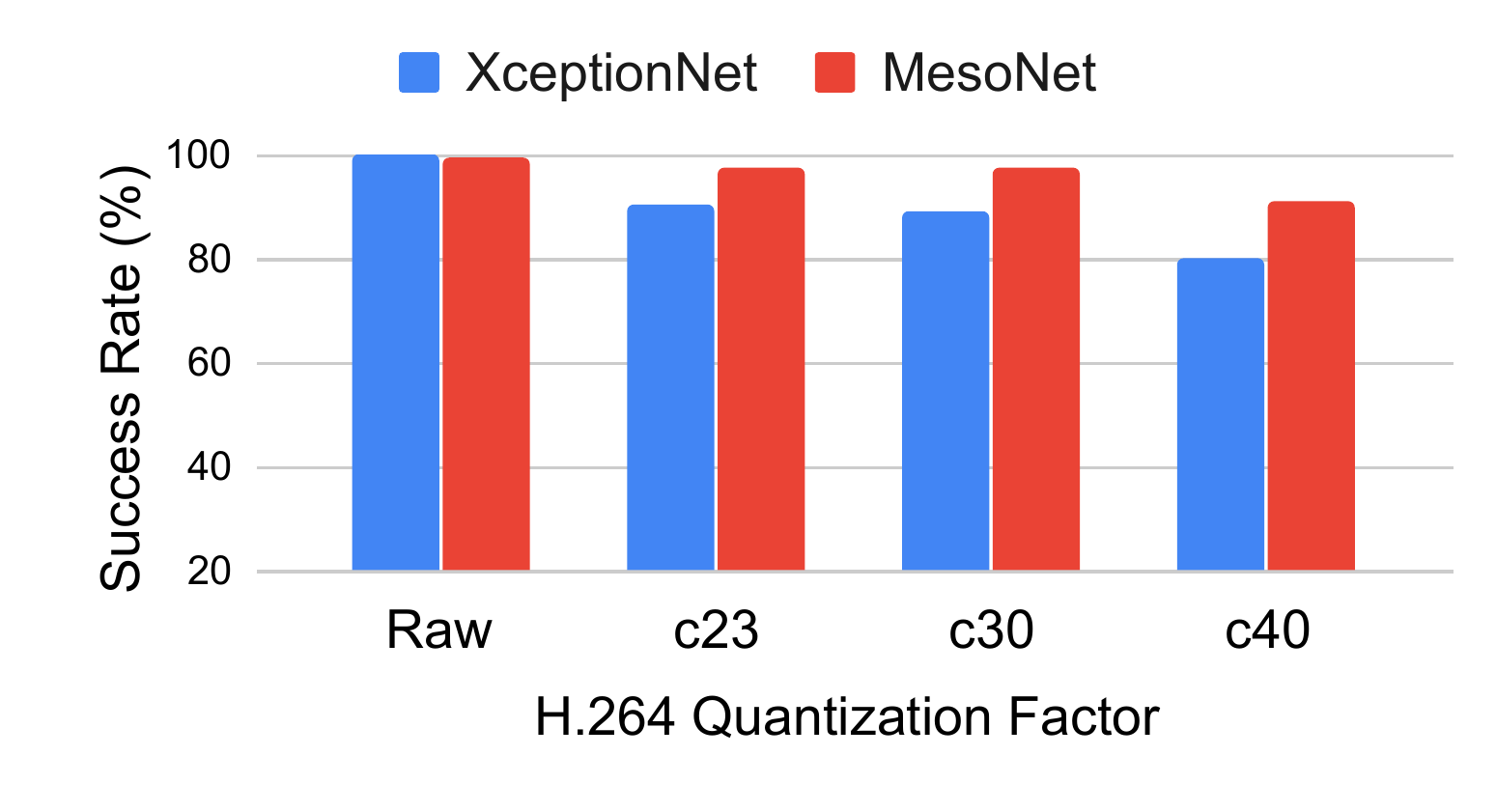}
    \vspace{-4mm}
    \caption{Attack success rate vs Quantization factor used for compression in H264 codec for robust white box attack. }
    \label{fig:wbh264}
    \vspace{-4mm}
\end{figure}

We also study the effectiveness of our robust white box attack under different levels of compression in the H.264 format which is widely used for sharing videos over the internet. Figure~\ref{fig:wbh264} shows the average success rate of our attack across all datasets for different quantization parameter $c$ used for saving the video in H.264 format. The higher the quantization factor, the higher is the compression level. In~\cite{faceforensicsiccv}, fake videos are saved in HQ and LQ formats which use $c=23$ and $c=40$ respectively. 
It can be seen that even at very high compression levels ($c=40$), our attack is able to achieve 80.39\% and 90.50\% attack success rate for XceptionNet and MesoNet respectively, without any additional hyper-parameter tuning for this experiment.

\subsection{Black-box Setting}
We construct adversarial examples in the black-box setting using the methods described in Sections \ref{sec:blackbox} and \ref{sec:blackbox2}.
The number of samples $n$ in the search distribution for estimating gradients using NES is set to 20 for black-box attacks and 80 for robust black-box to account for sampling different transformation functions $t_i$.  We set the maximum number of iterations to $100$, learning rate $\alpha$ to $1/255$ and max $L_\infty$ constraint $\epsilon$ to $16/255$. 

Table~\ref{tab:blackbox} shows the results of our Black-box attack (Section~\ref{sec:blackbox}) without robust transforms. Note that the average $L_\infty$ norm of the perturbation across all datasets and models is higher than our white-box attacks. We are able to generate adversarial videos with an average success rate of 97.04\% for XceptionNet and 86.70\% for MesoNet when adversarial videos are saved in the Raw format. Similar to our observation in the white-box setting, the success rate drops significantly in the compressed format for this attack. The average number of queries to the victim model for each frame is 985 for this attack.
\begin{table}[h]
\centering
\resizebox{\columnwidth}{!}{%
\begin{tabular}{@{}c|cccc|cccc@{}}
  \multicolumn{1}{c}{} & \multicolumn{4}{c}{\emph{XceptionNet}} & \multicolumn{4}{c}{\emph{MesoNet}} \\ \midrule
\textbf{Dataset} & $\mathbf{L_\infty}$ & \textbf{SR - U} & \textbf{SR - C} & \textbf{Acc-C\%} & $\mathbf{L_\infty}$ & \textbf{SR - U} & \textbf{SR - C} & \textbf{Acc-C \%} \\ \midrule
\textbf{DF}& 0.055 & 89.72 & 55.64 & 44.36 & 0.062 & 96.05 & 93.33 & 6.67   \\
\textbf{F2F}& 0.055 & 92.56 & 81.40 & 18.6 & 0.063 & 84.08 & 77.68 & 22.32 \\
\textbf{FS}& 0.045 & 96.77 & 23.50 & 76.5 & 0.063 & 77.55 & 62.44 & 37.56  \\
\textbf{NT}& 0.024 & 99.86 & 94.23 & 5.77 & 0.063 & 85.98 & 79.25 & 20.75 \\
\midrule
\textbf{All}& 0.045 & 94.73 & 63.69 & 36.31 & 0.063 & 85.92 & 78.18 & 21.82 \\
\bottomrule
\multicolumn{1}{c}{}
\end{tabular}%
}
\caption{Success Rate of of Black-box attack on XceptionNet and MesoNet. Acc-C denotes the accuracy of the detector on compressed adversarial videos. }
\label{tab:blackbox}
\end{table}

\noindent \textbf{Robust Black-box:} We perform robust black-box attack using the algorithm described in (Section \ref{sec:robustblackbox}).
For simplicity, during the robust black-box attack we use the same hyper-parameters for creating a distribution of transformation functions $T$ (Table \ref{tab:transforms}) as those in our robust white-box attack. 
The average number of network queries for fooling each frame is 2153 for our robust black-box attack.
Table \ref{tab:robblackbox} shows the results for our robust black-box attack.
We observe a significant improvement in the attack success rate for XceptionNet when we save adversarial videos in the compressed format as compared to that in the naive black-box attack setting. When attacking MesoNet in robust black-box setting, we do not observe a significant improvement even though overall success rate is higher when using robust transforms. 
\begin{table}[t]
\centering
\resizebox{\columnwidth}{!}{%
\begin{tabular}{@{}c|cccc|cccc@{}}
  \multicolumn{1}{c}{} & \multicolumn{4}{c}{\emph{XceptionNet}} & \multicolumn{4}{c}{\emph{MesoNet}} \\ \midrule
\textbf{Dataset} & $\mathbf{L_\infty}$ & \textbf{SR - U} & \textbf{SR - C} &  \textbf{Acc-C\%} & $\mathbf{L_\infty}$ & \textbf{SR - U} & \textbf{SR - C} & \textbf{Acc-C\%} \\ \midrule
\textbf{DF} & 0.060 & 88.47 & 79.18 & 20.82 & 0.047 & 96.19 & 93.80 & 6.20\\
\textbf{F2F} & 0.058 & 97.68 & 94.42 & 5.58 & 0.054 & 84.14 & 77.50 & 22.50\\
\textbf{FS} & 0.052 & 98.97 & 63.26 & 36.74 & 0.061 & 77.34 & 61.77 & 38.23 \\
\textbf{NT} & 0.018 & 99.65 & 98.91 & 1.09 & 0.053 & 88.05 & 80.27 & 19.73 \\
\midrule
\textbf{All} & 0.047 & 96.19 & 83.94 & 16.06 & 0.053 & 86.43 & 78.33 & 21.67 \\
\bottomrule
\multicolumn{1}{c}{}
\end{tabular}
}
\caption{Success Rate of Robust Black-box attack on XceptionNet and MesoNet. Acc-C denotes the accuracy of the detector on compressed adversarial videos. }
\label{tab:robblackbox}
\vspace{-6mm}
\end{table}
%-------------------------------------------------------------------------

\section{Evaluation on Sequence Based Detector}
\label{sec:temporalevaluation}
% In order 
We consider the 3D CNN based detector described in Section~\ref{sec:victimmodels}. The detector performs 3D convolution on a sequence of face-crops from 7 consecutive frames. We perform our attacks on the pre-trained model checkpoint (trained on DFDC~\cite{dolhansky2019deepfake} train set) released by the NTech-Lab team~\cite{sequencedetector}. We evaluate our attacks on the DeepFake videos from the DFDC public validation set which contains 200 Fake videos. We report the accuracy of the detector on the 7-frame sequences from this test set in the first row of Table~\ref{tab:temporalresults}.

Similar to our attacks on frame-by-frame detectors, in the white-box setting we back-propagate the loss through the entire model to obtain gradients with respect to the input frames for crafting the adversarial frames. While both white-box and robust white-box attacks achieve 100\% success rate on uncompressed videos, the robust white-box attack performs significantly better on the compressed videos and is able to completely fool the detector. As compared to frame-by-frame detectors, a higher magnitude of perturbation is required to fool this sequence model in both the white-box attacks.
In the black-box attack setting, while we achieve similar attack success rates on uncompressed videos as the frame-by-frame detectors, the attack success rate drops after compression. The robust black-box attack helps improve robustness of adversarial perturbations to compression as observed by higher success rates on compressed videos (51.02\% vs 24.43\% SR-C). 

\vspace{-2mm}
\setlength\tabcolsep{6pt} % default value: 6pt
\begin{table}[htbp]
\centering
\resizebox{0.9\columnwidth}{!}{%
\begin{tabular}{@{}l|cccc@{}}
  \multicolumn{1}{c}{} & \multicolumn{4}{c}{\emph{3D CNN Sequence Model}} \\ \midrule
\textbf{Attack Type} & $\mathbf{L_\infty}$ &
\textbf{SR - U} & \textbf{SR - C} & \textbf{Acc. - C\%} \\ \midrule
\textbf{None} & - & - & - & 91.74 \\
\midrule
\textbf{White-Box} & 0.037 & 100.00 & 77.67 & 22.33 \\
\textbf{Robust White-Box} & 0.059 & 100.00 & 100.00 & 0.00  \\
\textbf{Black-Box} & 0.061 & 87.99 & 24.43 & 75.57 \\
\textbf{Robust Black-Box} & 0.062 & 88.21 & 51.02 & 48.98\\
% \midrule
% \textbf{All} & 0.047 & 96.19 & 83.94 \\
\bottomrule
\multicolumn{1}{c}{}
\end{tabular}
}
\caption{Evaluation of different attacks on a sequence based detector on the DFDC validation dataset. The first row indicates the performance of the classifier on benign (non adversarial) videos. }
\label{tab:temporalresults}
\vspace{-5mm}
\end{table}

%-------------------------------------------------------------------------

\section{Discussion and Conclusion}
The intent of Deepfake generation can be malicious and their detection is a security concern.
Current works on DNN-based Deepfake detection assume a non-adaptive adversary whose aim is to fool the human-eye by generating a realistic fake video. 
To use these detectors in practice, we argue that it is essential to evaluate them against an adaptive adversary who is aware of the defense being present and is intentionally trying to fool the defense. In this paper, we show that the current state-of-the-art methods for Deepfake detection can be easily bypassed if the adversary has complete or even partial knowledge of the detector. 
Therefore, there is a need for developing provably robust detectors that are evaluated under different attack scenarios and attacker capabilities.

In order to use DNN based classifiers as detectors, ensuring robustness to adversarial examples is necessary but not sufficient. A well-equipped attacker may devise other methods to by-pass the detector: For example, an attacker can modify the training objective of the Deepfake generator to include a loss term corresponding to the detector score. Classifiers trained in a supervised manner on existing Deepfake generation methods, cannot be reliably secure against novel Deepfake generation methods not seen during training. 
We recommend approaches similar to Adversarial Training~\cite{goodfellow2014explaining} to train robust Deepfake detectors. That is, during training, an adaptive adversary continues to generate novel Deepfakes that can bypass the current state of the detector and the detector continues improving in order to detect the new Deepfakes. In conclusion, we highlight that adversarial examples are a practical concern for current neural network based Deepfake detectors and therefore recommend future work on designing provably robust Deepfake detectors.

\vspace{-2mm}
\section{Acknowledgements}
This work was supported by ARO under award number W911NF-19-1-0317 and SRC under Task ID: 2899.001.

{\small
\bibliographystyle{ieee_fullname}
\bibliography{egbib}
}

\end{document}